\documentclass[conference]{IEEEtran}
\IEEEoverridecommandlockouts
\usepackage{cite}
\usepackage{amsmath,amssymb,amsfonts}
\usepackage{algorithmic}
\usepackage{graphicx}
\usepackage{textcomp}
\usepackage{xcolor}
\usepackage{amsmath}
\usepackage{algorithm}
\usepackage{algorithmic}
\usepackage{booktabs}
\usepackage{multirow}
\usepackage{graphicx}
\usepackage{colortbl}
\usepackage{lipsum} 
\usepackage{amssymb}
\usepackage{makecell}
\usepackage{xcolor}
\usepackage{subcaption} 
\usepackage{pifont} 
\usepackage{arydshln}
\usepackage{newfloat}
\usepackage{listings}
\usepackage{verbatim}
\usepackage{hyperref}
\def\BibTeX{{\rm B\kern-.05em{\sc i\kern-.025em b}\kern-.08em
    T\kern-.1667em\lower.7ex\hbox{E}\kern-.125emX}}

\begin{document}

\title{IllusionBench+: A Large-scale and Comprehensive Benchmark for Visual Illusion Understanding in Vision-Language Models\\
\thanks{*Corresponding authors \\ 
This work was supported in part by the National Natural Science Foundation of China under Grant 62271312 and Grant 62132006, and in part by STCSM under Grant 22DZ2229005.}}

\author{Yiming Zhang, Zicheng Zhang, Xinyi Wei, Xiaohong Liu, Guangtao Zhai, Xiongkuo Min*\\
Institute of Image Communication and Network Engineering, \\
Shanghai Jiao Tong University, China\\
\{ming\_zhang\_sjtu, zzc1998, moj-will, xiaohongliu, zhaiguangtao, minxiongkuo\}@sjtu.edu.cn}
\vspace{0em}

\maketitle

\begin{abstract}
While current Visual Language Models (VLMs) exhibit remarkable image understanding capabilities, their responses to images with visual illusions, particularly those captured in real-world scenarios, remain underexplored. To address this gap, we construct \textbf{IllusionBench+}, a large-scale visual illusion dataset, encompassing 1,051 images, 5,548 question-answer pairs, and 1,051 golden text descriptions, covering the presence, causes, and content of illusions. We benchmark the performance of ten state-of-the-art (SOTA) VLMs on this dataset, evaluating their understanding of illusion images through true-or-false, multiple-choice, and open-ended descriptive tasks. The top-performing model, GPT-4o, achieves 80.59\% accuracy on true-or-false questions and 76.75\% accuracy on multiple-choice questions, but still lags behind human performance, indicating room for improvement in handling illusions. Furthermore, we observe that using the Step-by-step strategy, making the model first describe the presence and cause of the illusion before detailing the image content, improves the image understanding accuracy of most VLMs. To the best of our knowledge, IllusionBench+ is the largest and most comprehensive visual illusion benchmark for VLMs to date. The dataset is available in \url{https://github.com/mingZhang614/IllusionBench}
\end{abstract}


\section{Introduction}
\textbf{Visual illusions are perceptual anomalies caused by the visual system, characterized by a discrepancy between visual perception and reality} \cite{todorovic2020visual}. There are many types of illusions, and classifying them is challenging because the underlying causes are often unclear. However, Richard Gregory's classification \cite{gregory1991putting, gregory1997visual} provides a framework by dividing visual illusions into three main categories: physical illusions, physiological illusions, and cognitive illusions. Among these, cognitive visual illusions are the result of unconscious inferences and are perhaps the most widely recognized. Cognitive illusions are thought to arise from interactions with assumptions about the world, leading to what is termed ``unconscious inference" 
\cite{eagleman2011incognito}. Cognitive illusions are typically further categorized into ambiguous illusions, distortions, paradoxes, and fictions \cite{gregory1997visual, coren1976empirical}, which presents examples of each of these classic cognitive illusions. 
\begin{figure}[htbp]
    \centering
    \includegraphics[width=0.4  \textwidth]{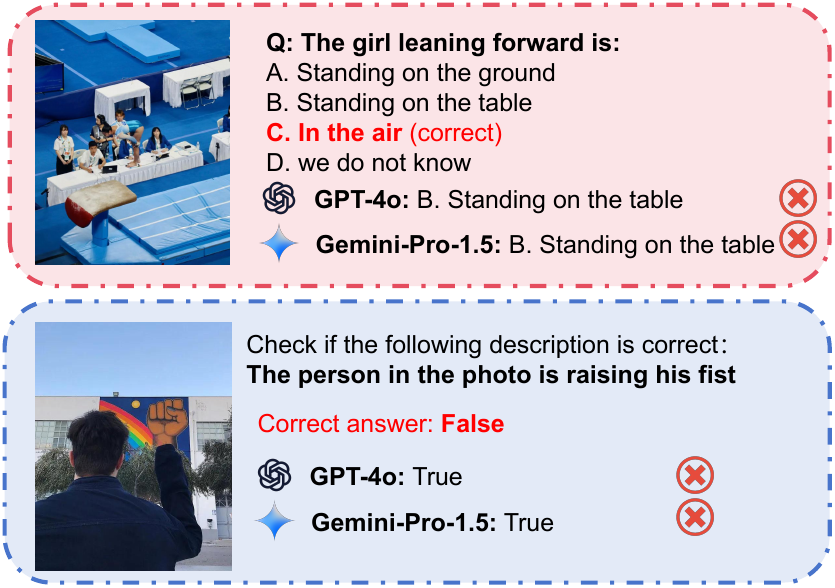} 
    \caption{Error cases from IllusionBench+.}
    \label{error_case}
\end{figure}
Although the mechanisms underlying visual illusions vary, these classic cognitive illusion images share a common feature: they are all artificially synthesized and inherently ambiguous.

In addition to artificially synthesized images, a small proportion of images captured in real-world scenes also exhibit visual illusions. The fundamental cause of this phenomenon is the inverse projection problem, where information is irreversibly lost during the projection from the three-dimensional world to two-dimensional images \cite{goldstein2022sensation}. This results in difficulties such as information loss, ambiguity, and multiple possible interpretations when attempting to infer three-dimensional objects and scenes from two-dimensional images (light and shadow projections)  \cite{palmer1999vision, gregory1997knowledge}.
\begin{figure}[htb]
    \centering
    \includegraphics[width=0.49\textwidth]{ 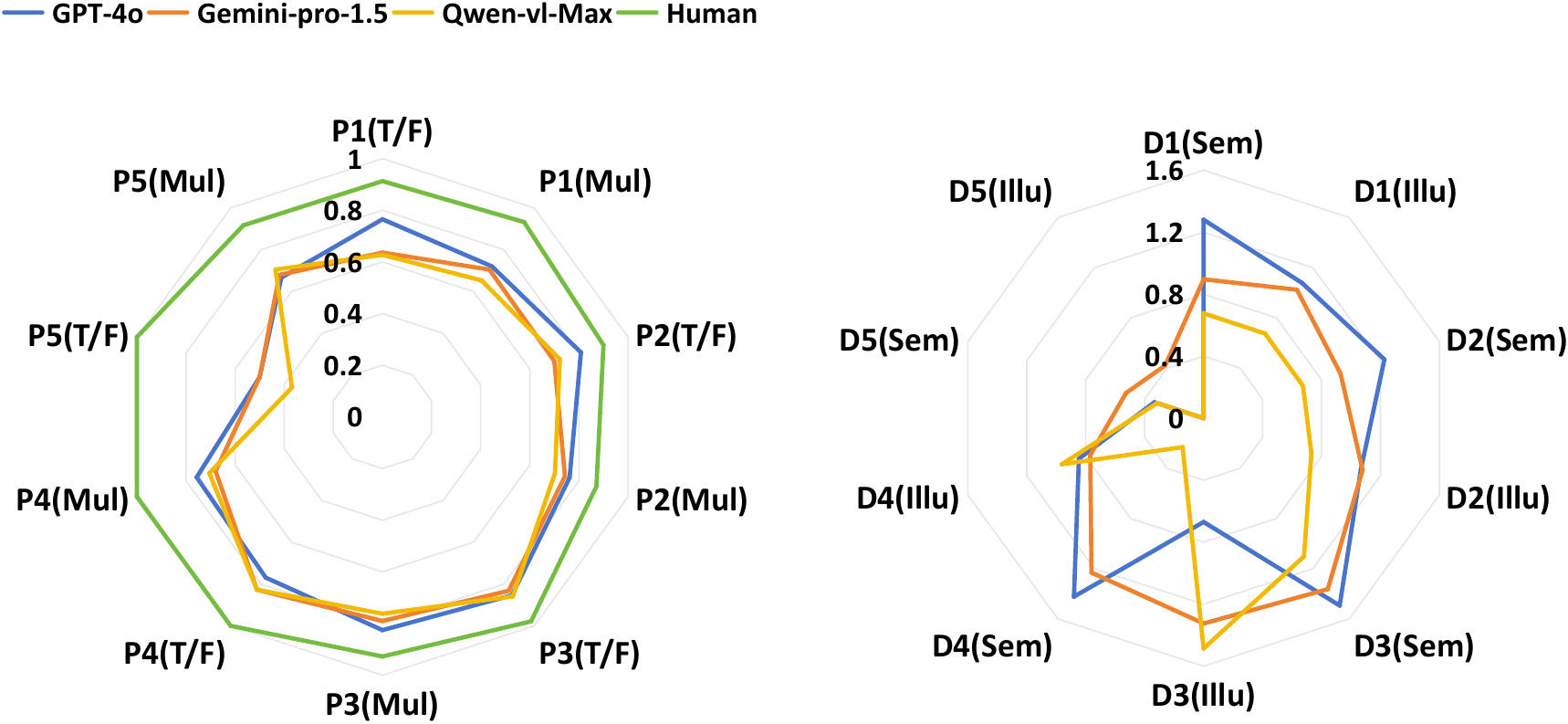} 
    \caption{Performance of advanced VLMs and human evaluators on IllusionBench+ perception tasks (left) and description tasks (right). The left image shows P1-P5 representing perception tasks on the subsets of \textbf{Classic Cognitive Illusion, Real Scene Illusion, No Illusion, Ishihara Image, and Trap Illusion}, respectively. Similarly, the right image shows D1-D5 representing description tasks on these subsets. ``T/F", ``Mul", ``Sem", and ``Illu" respectively represent true-or-false, multiple-choice, semantic descriptions, and illusion descriptions.}
    \label{radar}
\end{figure}

While the human brain compensates for the missing depth information in two-dimensional images (retinal projections) through binocular disparity and motion parallax  \cite{howard1995binocular, nakayama1994structure}, this issue remains unresolved in two-dimensional images captured by cameras. Consequently, both humans and vision models may experience visual illusions, leading to difficulties or errors in interpreting these images \cite{torralba2011unbiased, oliva2007role}.
To address this challenge, the human visual system leverages contextual cues for cognitive reasoning and utilizes monocular cues, such as perspective, occlusion relationships, shadows, and lighting, to alleviate the difficulties in information interpretation \cite{gregory1998eye, goldstein2022sensation2}. However, the extent to which current VLMs can recognize and interpret these visual illusions in real-world scenes remains an open question, as shown in Fig. \ref{error_case}.

\begin{figure*}[htb]
    \centering  
    \includegraphics[width= 0.75 \textwidth]{ 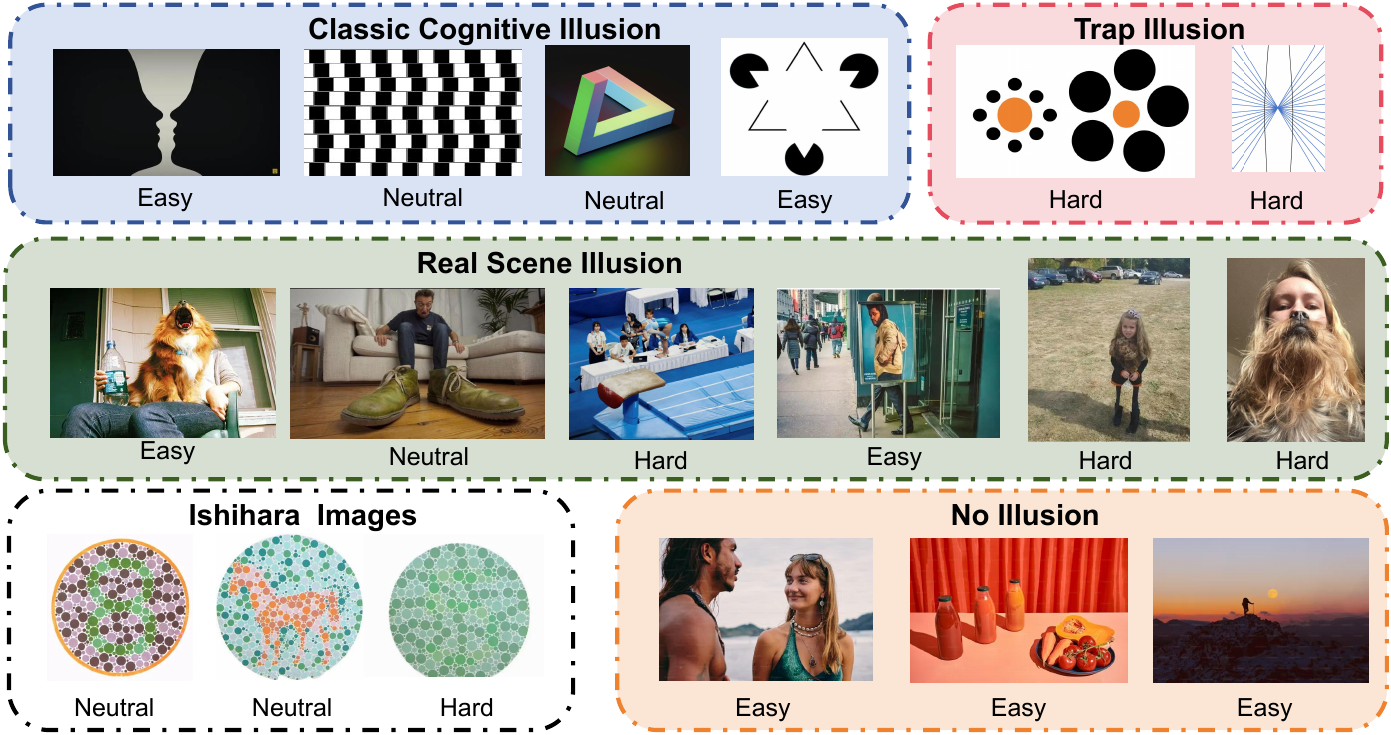}
    \caption{\textbf{Categories in IllusionBench+.} The annotations under each image represent the human cognitive difficulty score.}
    \label{Categories_fig}
\end{figure*}
Recent advancements in VLMs, like GPT-4o and Gemini-pro-1.5, have greatly improved visual question answering (VQA). These models, pre-trained on diverse datasets, excel in interpreting and responding to visual queries with increasing accuracy, handling tasks like object recognition, scene interpretation, and nuanced question answering \cite{10445007, Li_2024_CVPR, yin2023survey}. These improvements highlight their growing ability to bridge the gap between visual and textual information, enabling them to understand visual illusions.

\begin{table*}[htb]
\centering
\fontsize{8pt}{10pt}\selectfont
\caption{Comparison of IllusionBench+ with other illusion datasets}
\label{illusion_datasets}
\begin{tabular}{l ccccc}
\toprule
\textbf{Dataset} & \makecell{\textbf{Base}\\ \textbf{Image}} & \makecell{\textbf{Question}\\ \textbf{Type}} & \makecell{\textbf{Number of}\\ \textbf{Instance}} & \makecell{\textbf{Text}\\ \textbf{Description}} & \textbf{Image Type} \\
\midrule
GVIL & 16 & Binary & 1600 & × & Color \& Size illusions and variant \\
HallusionBench & 72 & Binary & 1129$^*$ & × & Color \& Size illusions and variant \\
\rule{0pt}{3ex}IllusionVQA\rule[-2ex]{0pt}{0pt} & 374 & Multiple-choice & 1435 & × & \makecell{12 types, mainly classical \\synthetic cognitive illusions} \\
\hdashline
\rule{0pt}{6ex}Ours\rule[-1ex]{0pt}{0pt} & 1051 & \makecell{Binary, \\Multiple-choice,\\Open-ended description} & 8716 & \checkmark & \makecell{Classic illusions, \\real scene illusions, \\trap illusions, no illusion,\\and Ishihara images} \\
\bottomrule
\end{tabular}\\
\footnotesize{$^*$ Note: The instances in HallusionBench include more than just visual illusions.}
\end{table*}

Previous research has used artificially synthesized classic cognitive visual illusion images as benchmarks for VLMs to explore the similarities between artificial intelligence and human visual cognition and to evaluate VLMs' understanding of visual illusions \cite{illusion1, illusion2, illusion3}. 
Unlike previous studies, our work includes not only classic cognitive illusion images, which lack real-world context, but also a large collection of real-scene visual illusions. These real-world images better represent practical applications and assess VLMs' ability to use contextual cues, similar to human perception. Additionally, because SOTA VLMs may have already learned classical illusions, these images may no longer be sufficient to test the visual perception ability of VLMs. To address the issue of potential overfitting to classic cognitive illusions, we introduce Ishihara color blindness detection images and trap illusion images. These images are accompanied by carefully crafted, manually annotated question-answer pairs, as well as image descriptions that cover image semantics, the presence of visual illusions, and their underlying causes.

Using our testing framework, we comprehensively evaluate the latest SOTA VLMs, such as GPT-4o, Gemini-pro-1.5, and several open-source models. Specifically, our framework includes true-or-false, multiple-choice, open-ended questions. We employ a step-by-step strategy, instructing the VLMs to describe the presence and cause of the illusion before describing the semantic content, which helps to alleviate the cognitive challenges that visual illusions pose to VLMs. Additionally, each image in our dataset is assigned a manually annotated cognitive difficulty level, and we conduct human testing to provide a multidimensional, fine-grained comparison between human performance and VLMs' performance on visual illusion cognition tasks. Fig. \ref{radar} shows the performances of SOTA VLMs on these tasks. 
Our contributions can be summarized as follows:

\begin{itemize}
    \item \textbf{IllusionBench+ Dataset:} We build a large-scale dataset that includes both classic and real-world visual illusions, color blindness test images, and trap illusions, supplemented with question-answer pairs and detailed annotations on image semantics, the presence of illusions, and their causes. To the best of our knowledge, IlluisionBench is the \textbf{largest} and most comprehensive visual illusion benchmark for VLMs to date.
    \item \textbf{Comprehensive Testing Framework:} We apply a rigorous framework to evaluate SOTA VLMs, such as GPT-4o and Gemini-pro-1.5, using a range of question types including true-or-false, multiple-choice, open-ended, ensuring a thorough evaluation of the models' capabilities in understanding visual illusions.
    \item \textbf{Step-by-step Strategy Insights:} Our Step-by-step strategy can enhance VLMs' semantic description performance when they accurately describe visual illusions, which can bring performance gains of up to 7.9\% and 33.7\% for GTP-4o and Gemini-pro-1.5, respectively. It demonstrates that accurately identifying visual illusions significantly improves models' semantic comprehension, highlighting the value of incorporating cognitive reasoning techniques to advance VLMs' understanding of complex visual information.
\end{itemize}

\section{Related Work}
\subsection{Visual Illusion and Visual Models}
Existing research has demonstrated that visual illusions for humans can induce equivalent illusions in models \cite{gomez2022synthesis, gomez2020color}. However, these studies primarily focus on specific types of illusions, including motion \cite{watanabe2018illusory}, brightness and color \cite{Gomez-Villa_2019_CVPR}, and completion \cite{kim2019neural}. Among them, \cite{gomez2022synthesis} proposed a framework that leverages the optimization capabilities of current automatic differentiation techniques to synthesize new visual illusions. This framework is used to study the differences between visual models and actual human perception, and it demonstrates that the illusions synthesized by the model should also be compelling to human observers.

\subsection{Benchmark on Illusion}
Recent studies have explored VLMs' ability to perceive visual illusions through natural language. A pioneering work \cite{illusion1} tested this by using a dataset of 1,600 variants from 16 root images, focusing on color and geometric distortions. The study aims to evaluate if SOTA VLMs align with human perception in visual illusions. Results show that while larger models perform better in localization tasks, VLMs generally struggle to interpret visual illusions as humans do.

Another study \cite{illusion2} introduced a benchmark to evaluate VLMs' handling of visual illusions and language hallucinations using a dataset of 346 images, including 72 focused on illusions, paired with question-answer tasks. The models, including GPT-4V, struggle with these illusions and hallucinations, achieving only 31.42\% accuracy. This highlights a misalignment with human perception and suggests that SOTA VLMs may overfit classic illusions, making them less effective for testing complex visual understanding.

Additionally, \cite{illusion3} introduced a dataset of 374 classic cognitive illusion images, generating 439 question-answer pairs to test VLMs' understanding and localization of challenging visual content. The study finds that advanced VLMs like GPT-4V and Gemini Pro perform poorly on visual illusions, with accuracy below that of human evaluators, highlighting current limitations in interpreting complex visual scenes.

Previous studies mainly focus on synthetic cognitive illusions, but our study expands this by including real-world scenes with visual illusions to better assess VLMs' use of contextual cues. Additionally, we introduce Ishihara and trap illusions to evaluate potential overfitting, ensuring a more precise alignment with human visual perception. To the best of our knowledge, IllusionBench+ is the largest and most comprehensive visual illusion benchmark for VLMs to date, as shown in Table \ref{illusion_datasets}.

\begin{figure}[htb]
    \centering
    \begin{subfigure}[b]{0.18\textwidth}
        \centering
        \includegraphics[width=\textwidth]{ 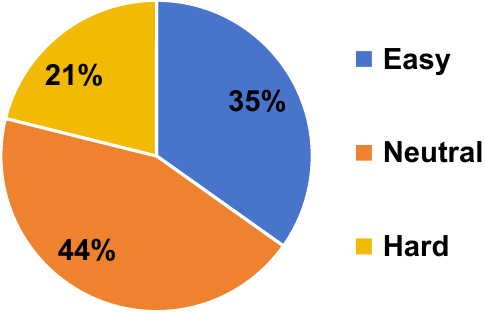} 
        \label{stat_a}
    \end{subfigure}
    \hfill
    \begin{subfigure}[b]{0.27\textwidth}
        \centering
        \includegraphics[width=\textwidth]{ 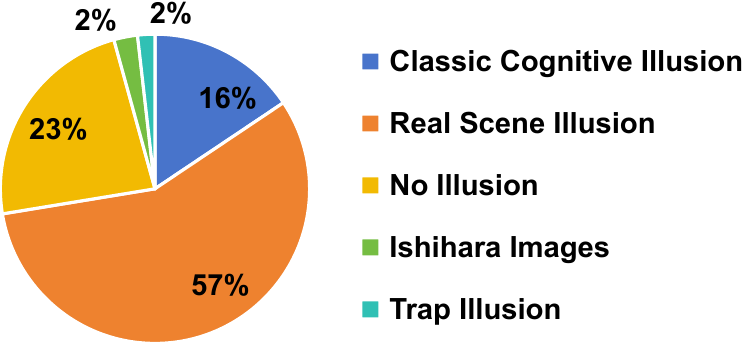} 
        \label{stat_b}
    \end{subfigure}
    
    \vskip\baselineskip
    
    \begin{subfigure}[b]{0.49\textwidth}
        \hspace*{0cm}
        \includegraphics[width= 0.8 \textwidth]{ 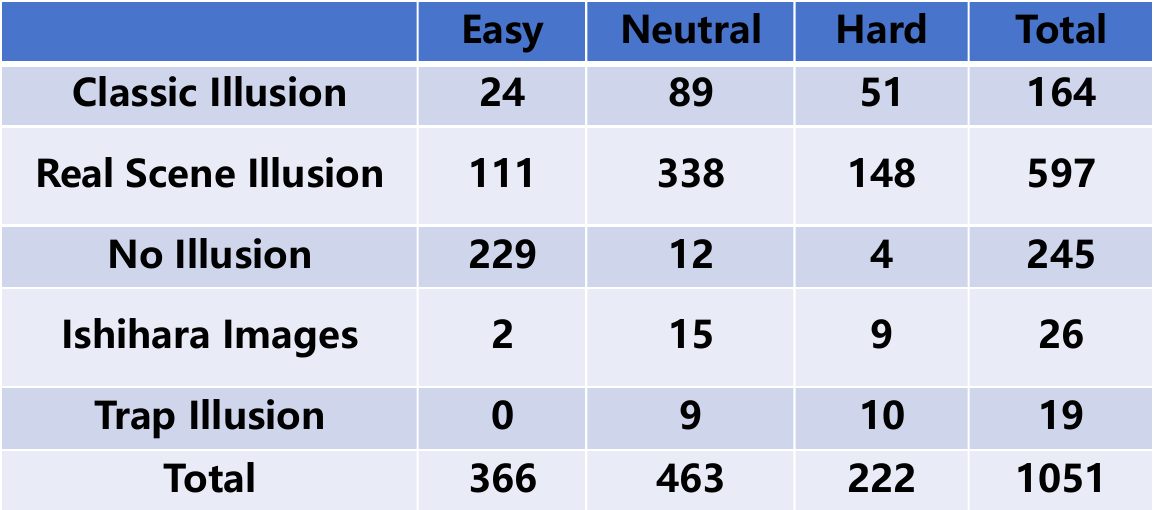}
        \label{stat_c}
    \end{subfigure}
    \caption{Dataset statistics. IllusionBench+ contains not only classic cognitive illusions but also a large number of real scene illusions, with real scenes accounting for more than 50\% of the total. It is also equipped with human cognitive difficulty scores, and it is the largest and most comprehensive visual illusion benchmark for VLM to date.}
    \label{component}
\end{figure}

\section{IllusionBench+}

\begin{figure*}[htb]
    \centering  
    \includegraphics[width= 0.75 \textwidth]{ 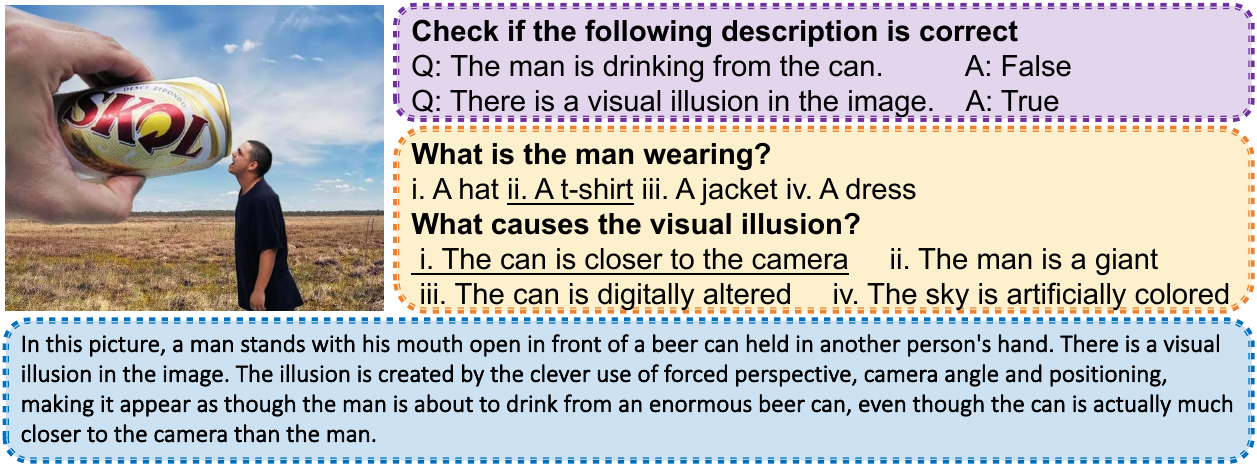}
    \caption{Example of real scene illusion in IllusionBench+. Each image in IllusionBench+ is equipped with at least two true-or-false questions, three multiple-choice questions, and a description that summarizes the semantic content of the image, the existence of visual illusions, and their causes.}
    \label{label_example}
\end{figure*}

To evaluate VLMs' understanding of visual illusions, we create IllusionBench+ with more than 1K images equipped with 5K QA pairs and manually annotated golden descriptions, as shown in Fig. \ref{label_example}. IllusionBench+ includes five image types: classical illusions, real scene illusions, no illusions, Ishihara images, and trap illusions (Fig. \ref{Categories_fig}). Testing tasks involve judgment, multiple-choice, and descriptive questions focused on illusion existence, causes, and content. This section details the dataset composition, question generation methods, and tasks.

\begin{table*}[htb]
\fontsize{8pt}{9pt}\selectfont
\belowrulesep=0pt
\aboverulesep=0pt
\centering
\caption{Performance of VLMs across different image categories and difficulty levels on IllusionBench+ true-or-false task. The best performance is marked in \textbf{bold}. ``Human" refers to the average performance of two human evaluators.}
\label{TF}
\begin{tabular}{l|ccccc|ccc|c}
\hline
Sub-category & \multicolumn{5}{c|}{\textbf{Image Category}} & \multicolumn{3}{c|}{\textbf{Difficulty Rating}} & \multirow{3}{*}{All} \\
\cmidrule(lr){1-1} \cmidrule(lr){2-6}  \cmidrule(lr){7-9} 
\multirow{2}{*}{VLMs} & Classical & Real Scene & No Illusion & Ishihara & Trap & \multirow{2}{*}{Easy} & \multirow{2}{*}{Neutral} & \multirow{2}{*}{Hard} & \\
& (P1) & (P2) & (P3) & (P4) & (P5) &&&& \\
\hline
\multicolumn{2}{l}{\textbf{Closed-Source VLMs}}\\
\hdashline
GPT-4o & \textbf{0.7653} & \textbf{0.8082} & 0.8532 & 0.7692 & 0.5000 & 0.8526 & 0.8040 & 0.7397 & \textbf{0.8059} \\
Gemini-pro-1.5 & 0.6363 & 0.6998 & 0.8319 & 0.8269 & 0.5000 & 0.7907 & 0.6943 & 0.6591 & 0.7183 \\
Qwen-vl-Max & 0.6276 & 0.7223 & \textbf{0.8589} & 0.8269 & 0.3684 & 0.8295 & 0.6913 & 0.6777 & 0.7338 \\
Qwen-vl-plus & 0.5522 & 0.6479 & 0.8250 & \textbf{0.9800} & 0.3055 & 0.7592 & 0.6447 & 0.6058 & 0.6742 \\
\hline
\multicolumn{2}{l}{\textbf{Opened-Source VLMs}}\\
\hdashline
CogVLM-17B (Vicuna-v1.5-7B) & 0.4028 & 0.4291 & 0.4431 & 0.4808 & \textbf{0.5263} & 0.4263 & 0.4352 & 0.4286  & 0.4308 \\
DeepSeek-VL-7B-chat & 0.3994 & 0.4705 & 0.4812 & 0.4694 & \textbf{0.5263} & 0.4682 & 0.4649 & 0.4478 & 0.4626 \\
InternLM-XComposer2-VL-7B (InternLM2) & 0.5552 & 0.6456 & 0.8033 & 0.7500 & 0.3158 & 0.7436 & 0.6379 & 0.5914 & 0.6625 \\
LLaVA-v1.5 (Vicuna-v1.5-7B) & 0.4128 & 0.4192 & 0.4741 & 0.4808 & \textbf{0.5263} & 0.4437 & 0.4319 & 0.4204  & 0.4333 \\
LLaVA-v1.5 (Vicuna-v1.5-13B) & 0.5145 & 0.7092 & 0.7847 & 0.7692 & 0.2632  & 0.7693 & 0.6719 & 0.6061 & 0.6895 \\           
LLaVA-NeXT (Llama3-8B) & 0.6221 & 0.6735 & 0.8302 & 0.7885 & \textbf{0.5263} & 0.7897 & 0.6586 & 0.6489 & 0.6995 \\
mPLUG-Owl2 (LLaMA-7B) & 0.5843 & 0.6154 & 0.7557 & 0.6731 & 0.3421 & 0.7001 & 0.6026 & 0.616& 0.6375 \\
Qwen-VL-Chat & 0.3866 & 0.4449 & 0.4534 & 0.4423 & \textbf{0.5263} & 0.4464 & 0.4416 & 0.4230 & 0.4391 \\
\hline
\textbf{\rule{0pt}{1.5ex}Human\rule[-1ex]{0pt}{0pt}} & 0.9130 & 0.9000 & 0.9787 & 1.0000 & 1.0000 & 0.9394 & 0.9170 & 0.9142 & 0.9234 \\
\hline
\end{tabular}
\end{table*}
\subsection{Collection and Composition of Images}
We collect 1K+ images from various online repositories. After manual selection, 780 images are confirmed to contain visual illusions, 26 are Ishihara color blindness detection images, and 245 imacges have no illusions, as shown in Fig. \ref{component}. The details are:
\begin{itemize}
    \item \textbf{Classic Cognitive Illusion Images }  
    These include blur, distortion, paradox, and fictitious illusions—key examples of traditional synthetic illusions. Designed by psychologists, these ambiguous images test VLMs' alignment with human perception. However, their classic nature and limited number may reduce their effectiveness, as they could be part of advanced VLMs' training datasets.
    \item \textbf{Trap Illusion } 
    Trap illusions are edited versions of classic visual illusions, resembling them in appearance but differing in physical properties. These images test whether VLMs overfit classic cognitive illusions, potentially causing hallucinations.
    \item \textbf{Real Scene Illusion Images } 
    IllusionBench+ includes 597 real-scene images with visual illusions. These images depict real-world objects and scenes, with unique and definite semantic descriptions. The illusions arise from the inverse projection problem, where information is lost in the transition from 3D to 2D. Understanding these images requires monocular cues like perspective, occlusion, shadows, and lighting, as well as contextual reasoning.
    \item \textbf{Ishihara Color Blindness Detection Images } IllusionBench+ includes 26 Ishihara images, verified by vision-healthy individuals, where the patterns convey unique and definite semantics. These images test whether VLMs' visual cognition aligns with human perception, specifically regarding Gestalt principles such as grouping, similarity, and proximity.
    \item \textbf{No Illusion Images } 
    IllusionBench+ contains 245 images with no illusions, depicting diverse subjects such as people, landscapes, and objects. These images provide a baseline for evaluating VLMs' visual understanding and the impact of illusions and evaluate the models' yes-bias when addressing questions about illusion presence.
\end{itemize}

\begin{table*}[htb]
\fontsize{8pt}{9pt}\selectfont
\belowrulesep=0pt
\aboverulesep=0pt
\centering
\caption{Performance of VLMs across different image categories and difficulty ratings on IllusionBench+ multiple-choice task. The best performance is marked in \textbf{bold}. ``Human" refers to the average performance of two human evaluators.}
\label{select}
\begin{tabular}{l|ccccc|ccc|c}
\hline
Sub-category & \multicolumn{5}{c|}{\textbf{Image Category}} & \multicolumn{3}{c|}{\textbf{Difficulty Rating}} & \multirow{3}{*}{All} \\
\cmidrule(lr){1-1} \cmidrule(lr){2-6}  \cmidrule(lr){7-9} 
\multirow{2}{*}{VLMs} & Classical & Real Scene & No Illusion & Ishihara & Trap & \multirow{2}{*}{Easy} & \multirow{2}{*}{Neutral} & \multirow{2}{*}{Hard} & \\
& (P1) & (P2) & (P3) & (P4) & (P5) &&&& \\
\hline
\multicolumn{2}{l}{\textbf{Closed-Source VLMs}}\\
\hdashline
GPT-4o & \textbf{0.7206} & \textbf{0.7620} & \textbf{0.8255} & \textbf{0.7564} & 0.6667 & \textbf{0.8163} & \textbf{0.7558} & \textbf{0.7172} & \textbf{0.7675}\\
Gemini-pro-1.5 & 0.7050 & 0.7432 & 0.7901 & 0.6795 & 0.6795 & 0.7998 & 0.7335 & 0.6818 & 0.7444 \\
Qwen-vl-Max & 0.6531 & 0.7026 & 0.7620 & 0.7051 & \textbf{0.7051} & 0.7608 & 0.6981 & 0.6392 & 0.7064 \\
Qwen-vl-plus & 0.5038 & 0.5715 & 0.7020 & 0.6410 & 0.5556 & 0.6720 & 0.5563 & 0.5339 & 0.5903\\
\hline
\multicolumn{2}{l}{\textbf{Opened-Source VLMs}}\\
\hdashline
CogVLM-17B (Vicuna-v1.5-7B) & 0.4624 & 0.4979 & 0.5943 & 0.5256 & 0.5256 & 0.5904 & 0.4980 & 0.4147 & 0.5112 \\
DeepSeek-VL-7B-chat & 0.3158 & 0.3603 & 0.3623 & 0.1538 & 0.1538 & 0.3843 & 0.3388 & 0.3078 & 0.3473 \\
InternLM-XComposer2-VL-7B (InternLM2) & 0.5188 & 0.6138 & 0.6961 & 0.4103 & 0.4103 & 0.6801 & 0.5863 & 0.5404 & 0.6077 \\
LLaVA-v1.5 (Vicuna-v1.5-7B) & 0.3195 & 0.3345 & 0.3835 & 0.2692 & 0.2692 & 0.3848 & 0.3134 & 0.3250 & 0.3395 \\
LLaVA-v1.5 (Vicuna-v1.5-13B) & 0.5075 & 0.5594 & 0.6274 & 0.4872 & 0.4872 & 0.6295 & 0.5322 & 0.5173 & 0.5612 \\
LLaVA-NeXT (Llama3-8B) & 0.5094 & 0.6042 & 0.6771 & 0.6410 & 0.6410 & 0.6515 &0.6013 & 0.5404 & 0.6050 \\
mPLUG-Owl2 (LLaMA-7B) & 0.4530 & 0.5137 & 0.5794 & 0.3333 & 0.3333 & 0.5621 & 0.4997 & 0.4540& 0.5107 \\
Qwen-VL-Chat & 0.3158 & 0.3614 & 0.3609 & 0.1538 & 0.1538 & 0.3834 & 0.3390 & 0.3106 & 0.3477 \\
\hline
\textbf{\rule{0pt}{1.5ex}Human\rule[-1ex]{0pt}{0pt}} & 0.9327 & 0.8712 & 0.9275 & 1.0000 & 0.9167 & 0.9170 & 0.8889 &0.8889 & 0.8975\\
\hline
\end{tabular}
\end{table*}

\subsection{Benchamrk on Illusion Perception Ability}
\subsubsection{Question Types and QA Pairs Generation}
The question-answer pairs in IllusionBench+ include both binary (true-or-false) and multiple-choice questions. Each image is accompanied by at least two binary questions and three multiple-choice questions, all manually annotated by humans. Each image also has a manually assigned cognitive difficulty rating, categorized as Easy, Neutral, or Hard, with all questions related to a given image sharing the same difficulty level.
\begin{itemize}
    \item \textbf{True-or-false Question:}  IllusionBench+ includes over 2,200 binary questions focused on semantic content and the presence of illusions, with 57\% of correct answers marked as False to counteract yes-bias in some VLMs. Semantic statements are intentionally misleading by human visual standards as shown in Fig. \ref{label_example}.
    
    \item \textbf{Multiple-choice Questions:} IllusionBench+ also features over 3,300 multiple-choice questions targeting fine-grained perception of image content and illusion causes. Each question offers four options, with one correct answer. Options are shuffled during evaluation.

\end{itemize}
\subsubsection{LLM-assisted Evaluation for VQA}
Our observations reveal that some VLMs do not output answers in the specified format. So we employ a \textit{LLM-Assisted Evaluation} method, which involves inputting the questions, correct answers, and VLM responses into a large language model (LLM) to evaluate the accuracy of the responses. Qwen-plus assisted in the evaluation of all models for 5 rounds.

While \textit{LLM-Assisted Evaluation} is efficient, it can sometimes err when the model’s output significantly deviates from the standard answer format. To address this, we manually review and correct all cases marked incorrect by the LLM. Thus, our evaluation combines manual and LLM-assisted methods for accuracy. Further details are in the Appendix. B.

\subsection{Benchmarks on Illusion Description Ability}
\subsubsection{Golden Description Definition and Question Type}
In addition to multiple question-answer pairs, each image is also accompanied by a manually crafted \textit{golden description}, covering the main content of the image, the existence of any visual illusion, and the causes of the illusion. The average length of each description is 53.21 words. All descriptions follow the format:

\textit{In this picture, [image semantics content]. There [is/is no] visual illusion in the image. The reason for visual illusions is [illusion causes].}

Supported by the golden descriptions, we conduct open-ended question-answer testing, which includes the following three types:

\begin{itemize}
    \item \textbf{Semantic Content Description }
    To evaluate whether VLMs can accurately describe the semantic content of the image with illusions, the prompt is:

    \textit{\# user: Please provide a description of the content in this image.}

    \item \textbf{Visual Illusion Description }
    To evaluate whether VLMs support the identification of visual illusions in the image and the accurate explanation of their cause, the prompt for this question type is:

    \textit{\# user: Is there a visual illusion in the image? If so, please explain the reason for the illusion.}

    \item \textbf{Step-by-Step Questioning }
    To test the impact of visual illusion cognition on semantic content understanding, VLMs are asked first to describe the visual illusion and then to describe the image's semantic content based on the memory of the previous answer. The experimental procedure is shown in Algorithm \ref{alg:visual_content_description}.
\end{itemize}

\begin{algorithm}[ht]
\caption{Step-by-Step Questioning with VLM}
\label{alg:visual_content_description}
\begin{algorithmic}[1]
\STATE \textbf{Input:} Image \( I \)
\STATE \textbf{Initialize:} Memory \(\text{memory} = \emptyset\)

\STATE \textbf{Step 1:} Ask VLM to describe the visual illusion. \textit{Prompt1:} \textit{Is there a visual illusion in the image? If so, please explain the reason for the illusion.}
\STATE \textbf{VLM Response:} \textit{response1}
\STATE \textbf{Update Memory:} \textit{memory} $\gets$ \textit{[Prompt1 + response1]}
\STATE \textbf{Step 2:} Describe the image's semantic content. \textit{Prompt2:} \textit{[memory] + ``Please provide a description of the content in this image.''}
\STATE \textbf{VLM Response:} \textit{response2}
\STATE \textbf{Output:} \textit{response2}

\end{algorithmic}
\end{algorithm}
\subsubsection{LLM-assisted Evaluation for Description}
This work examines how VLMs understand visual illusions, which often lead to challenges and inaccuracies in image interpretation. We evaluate VLM performance by assessing the accuracy of their descriptions, specifically their alignment with physical reality and human perception.

Previous studies have shown that single-modal language models are effective for evaluating language tasks \cite{zheng2024judging}. After collecting open-ended responses from the VLMs, we use advanced LLMs to quantitatively evaluate multimodal description tasks. Specifically, both the model’s output and the golden description are input into the LLM, which compares the two to identify significant conflicts. Preciseness is scored on a scale of [0, 1, 2]. 
The evaluations of all models are assisted by Qwen-plus for 5 rounds. Our human study shows that Spearman's rank correlation coefficient (SRCC) between LLM and human evaluation results exceeds 0.9. Details regarding prompts and other specifics can be found in the Appendix. C.

\section{Experiment Setup}
\subsection{Vision Language Models}
We test four SOTA closed-source models and eight open-source models. The closed-source models include GPT-4o (version 2024-05-13) \cite{achiam2023gpt}, Gemini-pro-1.5 (latest update in May 2024) \cite{reid2024gemini}, Qwen-VL-Plus, and Qwen-VL-Max \cite{qwen-vl}. We use the latest versions available at the time of writing, with their default API parameters. The open-source models include CogVLM-17B (Vicuna-v1.5-7B) \cite{cogvlm}, DeepSeek-VL-7B-chat \cite{deepseek}, InternLM-XComposer2-VL-7B (InternLM2) \cite{internlm}, LLaVA-v1.5 (Vicuna-v1.5-7B), LLaVA-v1.5 (Vicuna-v1.5-13B), LLaVA-NeXT (Llama3-8B) \cite{llava_next}, mPLUG-Owl2 (LLaMA-7B) \cite{mplug}, and Qwen-VL-Chat \cite{qwen-vl}. These models span different architectures and parameter scales, are trained on a wide range of vision-language tasks, and exhibit strong visual understanding capabilities.

\subsection{Human vs VLMs}
To evaluate the alignment between VLMs' perception of visual illusion images and human visual perception, we utilize a subset of IllusionBench+ to evaluate human visual illusion perception. We recruited two human evaluators and provided them with a subset of 200 sampled images from the dataset, proportionally sampled according to image categories. The human evaluators completed all multiple-choice and judgment questions within this subset. We then quantify human cognitive abilities using the same LLM-assisted method described earlier.

\begin{table*}[htb]
\centering
\fontsize{8pt}{9.2pt}\selectfont
\caption{Performance of VLMs on IllusionBench+ description task. The best performance is marked in \textbf{bold}. The blue part represents the standard deviation between samples.}
\label{description1}
\begin{tabular}{lcccc}
\toprule
Sub-category & \multicolumn{3}{c}{\textbf{Open-ended Question Type}} &  \multirow{3}{*}{All} \\
\cmidrule(lr){1-1} \cmidrule(lr){2-4} 
VLMs & \makecell{Semantic Content\\Description} &\makecell{Visual Illusion\\Description}& \makecell{Step-by-Step\\Questioning} & \\
\hline
\textbf{Closed-Source VLMs} &&&\\
\hdashline
GPT-4o & 
\textbf{1.2872} \textcolor{cyan}{$\pm$} \textcolor{cyan}{0.9315} & 0.9522 \textcolor{cyan}{$\pm$} \textcolor{cyan}{0.7546} & \textbf{1.3616} \textcolor{cyan}{$\pm$} \textcolor{cyan}{0.8989} & \textbf{1.2004} \textcolor{cyan}{$\pm$} \textcolor{cyan}{0.8830}\\
Gemini-pro-1.5 & 
1.0257 \textcolor{cyan}{$\pm$} \textcolor{cyan}{0.9789} & \textbf{1.1077} \textcolor{cyan}{$\pm$} \textcolor{cyan}{0.7917} & 1.1954 \textcolor{cyan}{$\pm$} \textcolor{cyan}{0.9556} & 1.1096 \textcolor{cyan}{$\pm$} \textcolor{cyan}{0.9149} \\
Qwen-vl-Max & 
0.7571 \textcolor{cyan}{$\pm$} \textcolor{cyan}{0.9492} & 0.8913 \textcolor{cyan}{$\pm$} \textcolor{cyan}{0.7652} & 0.9038 \textcolor{cyan}{$\pm$} \textcolor{cyan}{0.9672} & 0.8507 \textcolor{cyan}{$\pm$} \textcolor{cyan}{0.9007}  \\
Qwen-vl-plus & 
0.7924 \textcolor{cyan}{$\pm$} \textcolor{cyan}{0.9490} & 0.7052 \textcolor{cyan}{$\pm$} \textcolor{cyan}{0.6840} & 0.8029 \textcolor{cyan}{$\pm$} \textcolor{cyan}{0.9437} & 0.7668 \textcolor{cyan}{$\pm$} \textcolor{cyan}{0.8687} \\
\hline
\textbf{Opened-Source VLMs} &&&\\
\hdashline
CogVLM-17B (Vicuna-v1.5-7B) & 
0.9001 \textcolor{cyan}{$\pm$} \textcolor{cyan}{0.9703} & 0.7222 \textcolor{cyan}{$\pm$} \textcolor{cyan}{0.7476} & 0.8535 \textcolor{cyan}{$\pm$} \textcolor{cyan}{0.9658} & 0.8252 \textcolor{cyan}{$\pm$} \textcolor{cyan}{0.9034} \\
DeepSeek-VL-7B-chat & 
0.7550 \textcolor{cyan}{$\pm$} \textcolor{cyan}{0.9518} & 0.7293 \textcolor{cyan}{$\pm$} \textcolor{cyan}{0.7576} & 0.8071 \textcolor{cyan}{$\pm$} \textcolor{cyan}{0.9559} & 0.7636 \textcolor{cyan}{$\pm$} \textcolor{cyan}{0.8940} \\
InternLM-XComposer2-VL-7B (InternLM2) & 
0.7431 \textcolor{cyan}{$\pm$} \textcolor{cyan}{0.9313} & 0.6552 \textcolor{cyan}{$\pm$} \textcolor{cyan}{0.6981} & 0.7336 \textcolor{cyan}{$\pm$} \textcolor{cyan}{0.9337} & 0.7107 \textcolor{cyan}{$\pm$} \textcolor{cyan}{0.8621} \\
LLaVA-v1.5 (Vicuna-v1.5-7B) & 
0.4814 \textcolor{cyan}{$\pm$} \textcolor{cyan}{0.8260} & 0.3333 \textcolor{cyan}{$\pm$} \textcolor{cyan}{0.4971} & 0.4177 \textcolor{cyan}{$\pm$} \textcolor{cyan}{0.7781} & 0.4107 \textcolor{cyan}{$\pm$} \textcolor{cyan}{0.7176} \\
LLaVA-v1.5 (Vicuna-v1.5-13B) & 
0.5290 \textcolor{cyan}{$\pm$} \textcolor{cyan}{0.8540} & 0.3626 \textcolor{cyan}{$\pm$} \textcolor{cyan}{0.5228} & 0.3777 \textcolor{cyan}{$\pm$} \textcolor{cyan}{0.7541} & 0.4232 \textcolor{cyan}{$\pm$} \textcolor{cyan}{0.7276}  \\
LLaVA-NeXT (Llama3-8B) & 
0.7364 \textcolor{cyan}{$\pm$} \textcolor{cyan}{0.9431} & 0.7165 \textcolor{cyan}{$\pm$} \textcolor{cyan}{0.7339} & 0.8192 \textcolor{cyan}{$\pm$} \textcolor{cyan}{0.9659} & 0.7574 \textcolor{cyan}{$\pm$} \textcolor{cyan}{0.8880}  \\
mPLUG-Owl2 (LLaMA-7B) & 
0.5975 \textcolor{cyan}{$\pm$} \textcolor{cyan}{0.8916} & 0.3840 \textcolor{cyan}{$\pm$} \textcolor{cyan}{0.5458} & 0.5385 \textcolor{cyan}{$\pm$} \textcolor{cyan}{0.8681} & 0.5068 \textcolor{cyan}{$\pm$} \textcolor{cyan}{0.7897}  \\
Qwen-VL-Chat & 
0.7336 \textcolor{cyan}{$\pm$} \textcolor{cyan}{0.9408} & 0.5891 \textcolor{cyan}{$\pm$} \textcolor{cyan}{0.6962} & 0.7488 \textcolor{cyan}{$\pm$} \textcolor{cyan}{0.9480} & 0.6906 \textcolor{cyan}{$\pm$} \textcolor{cyan}{0.8724}  \\

\bottomrule
\end{tabular}
\end{table*}

\definecolor{mycolor}{RGB}{100,193,80}

\begin{table*}[htb]
\centering
\fontsize{8pt}{9.2pt}\selectfont
\caption{Performance of VLMs on semantic content description task with and without Step-by-step strategy under different illusion description preciseness level. Red represents the performance improvement caused by the step-by-step strategy, while green represents the performance decline.}
\label{description2}
\begin{tabular}{lccccccc}
\toprule

\multirow{2}{*}{Sub-category} & \multicolumn{2}{c}{Illusion Preciseness} & \multicolumn{2}{c}{Illusion Preciseness} & \multicolumn{2}{c}{Illusion Preciseness} &  \\

& \multicolumn{2}{c}{\textbf{$\in [1.33, 2]$}} & \multicolumn{2}{c}{\textbf{$\in [0.67, 1.33)$}} & \multicolumn{2}{c}{\textbf{$\in [0, 0.67)$}} &  \\

\cmidrule(lr){1-1}  \cmidrule(lr){2-7} 
VLMs &Semantic & Step-by-Step & Semantic &Step-by-Step & Semantic & Step-by-Step  \\
\hline
\textbf{\rule{0pt}{2ex}Closed-Source VLMs\rule[-1ex]{0pt}{0pt}} &&&&&&\\
\hdashline
GPT-4o & 1.4745 & 1.5146  (\textcolor{red}{$\uparrow$}\textcolor{red}{2.7\%}) & 1.2138 & 1.3096 (\textcolor{red}{$\uparrow$}\textcolor{red}{7.9\%}) & 1.2346 & 1.3086 (\textcolor{red}{$\uparrow$}\textcolor{red}{6.0\%}) \\
Gemini-pro-1.5 & 1.3529 & 1.5627 (\textcolor{red}{$\uparrow$}\textcolor{red}{15.5\%})  & 0.8816 & 1.1789  (\textcolor{red}{$\uparrow$}\textcolor{red}{33.7\%}) & 0.7626 & 0.7014  (\textcolor{mycolor}{$\downarrow$}\textcolor{mycolor}{8.0\%}) \\
Qwen-vl-Max & 1.0391 & 1.1875 (\textcolor{red}{$\uparrow$}\textcolor{red}{14.3\%}) & 0.7329 & 0.9953  (\textcolor{red}{$\uparrow$}\textcolor{red}{35.8\%}) & 0.5919 & 0.6054  (\textcolor{red}{$\uparrow$}\textcolor{red}{2.3\%}) \\
Qwen-vl-plus & 0.9412 & 0.9412 (\textcolor{black}{$\uparrow$}\textcolor{black}{0.0\%}) & 0.7388 & 0.8180 (\textcolor{red}{$\uparrow$}\textcolor{red}{10.7\%}) & 0.8067 & 0.7483 (\textcolor{mycolor}{$\downarrow$}\textcolor{mycolor}{7.2\%})\\
\hline
\textbf{\rule{0pt}{2.5ex}Opened-Source VLMs\rule[-1ex]{0pt}{0pt}} &&&&&&\\
\hdashline
CogVLM-17B (Vicuna-v1.5-7B) & 1.2766 & 1.2979  (\textcolor{red}{$\uparrow$}\textcolor{red}{1.7\%}) & 0.9739 & 1.0705  (\textcolor{red}{$\uparrow$}\textcolor{red}{9.9\%}) & 0.6938 & 0.5062  (\textcolor{mycolor}{$\downarrow$}\textcolor{mycolor}{27.0\%}) \\
DeepSeek-VL-7B-chat & 1.0879 & 0.9779  (\textcolor{mycolor}{$\downarrow$}\textcolor{mycolor}{10.1\%})& 0.8041 & 1.0760  (\textcolor{red}{$\uparrow$}\textcolor{red}{33.8\%})& 0.6171 & 0.5252 (\textcolor{mycolor}{$\downarrow$}\textcolor{mycolor}{14.9\%})  \\
InternLM-XComposer2-VL-7B (InternLM2) & 1.1460 & 1.1460 (\textcolor{black}{$\uparrow$}\textcolor{black}{0.0\%}) & 0.7319 & 0.6957 (\textcolor{mycolor}{$\downarrow$}\textcolor{mycolor}{4.9\%}) & 0.6433 & 0.6533  (\textcolor{red}{$\uparrow$}\textcolor{red}{1.6\%})  \\
LLaVA-v1.5 (Vicuna-v1.5-7B) & 1.3846 & 1.5385 (\textcolor{red}{$\uparrow$}\textcolor{red}{11.1\%}) & 0.6142 & 0.7623 (\textcolor{red}{$\uparrow$}\textcolor{red}{24.1\%}) & 0.4048 & 0.2409  (\textcolor{mycolor}{$\downarrow$}\textcolor{mycolor}{40.5\%}) \\
LLaVA-v1.5 (Vicuna-v1.5-13B) & 1.0000 & 0.9545 (\textcolor{mycolor}{$\downarrow$}\textcolor{mycolor}{4.6\%}) & 0.4405 & 0.4375  (\textcolor{mycolor}{$\downarrow$}\textcolor{mycolor}{0.7\%})& 0.5594 & 0.3319  (\textcolor{mycolor}{$\downarrow$}\textcolor{mycolor}{40.7\%})\\
LLaVA-NeXT (Llama3-8B) & 1.2784 & 1.3409 (\textcolor{red}{$\uparrow$}\textcolor{red}{4.9\%}) & 0.7282 & 1.2195 (\textcolor{red}{$\uparrow$}\textcolor{red}{67.5\%}) & 0.5422 & 0.2869  (\textcolor{mycolor}{$\downarrow$}\textcolor{mycolor}{47.1\%})\\
mPLUG-Owl2 (LLaMA-7B) & 0.7500 & 0.8125 (\textcolor{red}{$\uparrow$}\textcolor{red}{8.3\%}) & 0.5976 & 0.6953 (\textcolor{red}{$\uparrow$}\textcolor{red}{16.3\%}) & 0.5923 & 0.4505 (\textcolor{mycolor}{$\downarrow$}\textcolor{mycolor}{23.9\%})  \\
Qwen-VL-Chat & 1.1732 & 1.2441  (\textcolor{red}{$\uparrow$}\textcolor{red}{6.0\%})& 0.8654 & 1.0110  (\textcolor{red}{$\uparrow$}\textcolor{red}{16.8\%})& 0.5466 & 0.4677  (\textcolor{mycolor}{$\downarrow$}\textcolor{mycolor}{14.4\%})  \\
\bottomrule
\end{tabular}
\end{table*}

\section{Result on IllusionBench+}
\subsection{Result on Illusion Perception}
\textbf{The existence of visual illusions significantly affects the visual perception of VLMs.}
We evaluate VLMs' ability to perceive visual illusions using true-or-false and multiple-choice tasks, with results in Table \ref{TF} and Table \ref{select}, revealing several key insights:

1) GPT-4o performs best in both tasks, with a true-or-false accuracy of 0.8059 and multiple-choice accuracy of 0.7675, but still lags behind human performance, indicating room for improvement in handling illusions.

2) Performances of all VLMs vary across image categories, with higher accuracy for no-illusion images and real-scene illusions compared to classical cognition illusions. GPT-4o excels in classic illusions but underperforms in trap illusions, likely due to hallucinations when encountering patterns similar to classic ones, suggesting that testing VLMs with only classic illusions is insufficient.

3) We also use Ishihara color blindness test images to examine if VLMs' perception aligns with Gestalt principles. Qwen-vl-plus shows the highest judgment accuracy (0.98), nearing the human level, but the multiple-choice performance is weaker, highlighting gaps in fine-grained perception and specific knowledge of the Ishihara test. Other VLMs all have gaps with humans in both tasks.

\subsection{Result on Illusion Description}
\textbf{Accurately describing visual illusions significantly enhances the visual perception capabilities of VLMs.}
The performance results for VLMs on the open-ended description task are shown in Table \ref{description1} and Table \ref{description2}, revealing several key insights: 

1) As shown in Table \ref{description1}, GPT-4o achieves the highest overall performance in the description task. The open-source model CogVLM-17B performs comparably to the closed-source Qwen-vl series.

2) We adopt a Step-by-step strategy, first asking about the presence and cause of the illusion before inquiring about the image content. As shown in Table \ref{description1}, this strategy improves semantic accuracy for all closed-source VLMs, but not for the open-source models. The result indicate that models benefiting from the step-by-step approach typically have higher visual illusion description performance.

3) The step-by-step strategy's effectiveness depends on VLMs' ability to describe visual illusions, as shown in Table \ref{description2}. When VLMs accurately recognize illusions, semantic performance improves, especially at moderate accuracy levels. This indicates that most VLMs have some understanding of illusions, and the step-by-step strategy effectively leverages this potential when they encounter mild comprehension difficulties. However, if VLMs struggle with illusion recognition, performance may decline. These findings highlight the importance of improving VLMs' perception of visual illusions to enhance their overall visual understanding.

\section{Conclusion}
In this study, we introduce IllusionBench+, the most extensive and comprehensive benchmark for evaluating VLMs on visual illusions. Our findings demonstrate that while SOTA VLMs, like GPT-4o, perform well in various tasks, they still struggle to interpret visual illusions accurately, highlighting a significant gap between model performance and human perception. The step-by-step strategy we employed shows the potential to enhance the models' semantic understanding when they accurately describe visual illusions. However, the persistent challenges indicate that there is still much room for improvement in aligning VLMs with human visual cognition. IllusionBench+ can bring VLMs closer to human-like understanding and interpretation of complex visual scenes.

Future work will focus on expanding the diversity of illusion types within IllusionBench+. Additionally, we aim to explore more sophisticated evaluation strategies and model architectures that could better mimic human cognitive processes in visual perception tasks. This continued research will contribute to advancing the capabilities of VLMs, bringing them closer to human-like understanding and interpretation of complex visual scenes.

\newpage
\bibliographystyle{IEEEbib}
\bibliography{references}
\clearpage

\appendix
\subsection{Appendix A: More Information on Illusion Perception Tasks}
\subsubsection{QA Pairs Generation and Prompts}
The question-answer pairs in IllusionBench+ include both binary (true-or-false) and multiple-choice questions. Each image is accompanied by at least two binary questions and three multiple-choice questions, all manually annotated by humans. Each image also has a manually assigned cognitive difficulty rating, categorized as Easy, Neutral, or Hard, with all questions related to a given image sharing the same difficulty level.
\begin{itemize}
    \item \textbf{True-or-False Question:}  IllusionBench+ contains over 2,200 binary questions, primarily addressing semantic content and the presence of illusions. The semantic statements are designed to be the most misleading according to human visual standards. Notably, trap illusions are labeled as containing illusions, while Ishihara color blindness detection images are labeled as not containing illusions. Due to the tendency of some VLMs to exhibit yes-bias, where they prefer to answer true/yes, 57\% of the binary questions in IllusionBench+ have False as the correct answer. The input format for the question is as follows:
    
    \textit{\# user: [Image Tokens] Given an image, check if the following description is correct, and answer `True' or `False'. Do not explain the reason. Description: [True-or-False Question]}
    
    \item \textbf{Multiple-choice Questions:} In addition to True-or-False questions, IllusionBench+ also includes over 3,300 multiple-choice questions, focusing on the fine-grained perception of image content and the specific causes of illusions. Each question offers four carefully designed options, with only one being correct. The correct and wrong answers are shuffled during the actual evaluation. The input format for the question is as follows:
    
    \textit{\# user: [Image Tokens] Given an image, a question, and some options, You have to select the correct one. Do not explain your reasoning. Answer with only the Roma number that corresponds to the correct option. Do not repeat the entire answer. Do not explain the reason. [Multiple-choice Questions and Options]}
    
\end{itemize}
\subsubsection{True/False Bias Test}

Based on our observations, some models tend to favor either true or false responses when completing true-or-false tasks. Therefore, we further analyzed this bias in the models' answers. We used the False Positive Ratio (FP Ratio) to characterize the models' response preferences and quantify the tendency of the VLMs to incorrectly classify a negative instance as positive. It is defined as the proportion of false positives relative to all incorrect predictions. Mathematically, the FP Ratio can be expressed as:
$$ {R_{FP}} = \frac{FP}{FP + FN}$$
where \(FP\) represents the number of instances where the VLM incorrectly answers ``True" when the correct answer is ``False". \(FN\) represents the number of instances where the VLM incorrectly answers ``False" when the correct answer is ``True". \(FP +FN\) denotes the total number of incorrect answers made by the VLM. The $R_{FP}$ closer to 0.5 indicates greater robustness in the VLM, while a ratio near 1 suggests a bias toward answering ``True", and a ratio near 0 indicates a bias towards answering ``False". 

As shown in Table. \ref{TF_FP}, CogVLM, DeepSeek-VL-7B-chat, LLaVA-v1.5 (Vicuna-v1.5-7B), and Qwen-VL-Chat exhibit a strong tendency to answer ``True", which results in lower performance for these VLMs on true-or-false tasks. One possible reason is that the training datasets for these models contain a majority of samples labeled as ``True", with insufficient instances of negative responses. This imbalance may lead the models to learn this bias, causing them to favor ``True" answers in judgment tasks.

\begin{table*}[htb]
\fontsize{8pt}{9pt}\selectfont
\belowrulesep=0pt
\aboverulesep=0pt
\centering
\caption{FP ratio of VLMs across different image categories and difficulty levels on IllusionBench+ true-or-false task.}
\label{TF_FP}
\begin{tabular}{l|ccccc|ccc|c}
\hline
Sub-category & \multicolumn{5}{c|}{\textbf{Image Category}} & \multicolumn{3}{c|}{\textbf{Difficulty Rating}} & \multirow{3}{*}{All} \\
\cmidrule(lr){1-1} \cmidrule(lr){2-6}  \cmidrule(lr){7-9} 
\multirow{2}{*}{VLMs} & Classical & Real Scene & No Illusion & Ishihara & Trap & \multirow{2}{*}{Easy} & \multirow{2}{*}{Neutral} & \multirow{2}{*}{Hard} & \\
& (P1) & (P2) & (P3) & (P4) & (P5) &&&& \\
\hline
\multicolumn{2}{l}{\textbf{Closed-Source VLMs}}\\
\hdashline
GPT-4o & 0.4 & 0.372 & 0.3571 & 0.3571 & 0.3571 & 0.3364 & 0.3571 & 0.4286 & 0.373 \\
Gemini-pro-1.5 & 0.6341 & 0.6253 & 0.4937 & 0.4937 & 0.4937 & 0.5563 & 0.6060 & 0.6280 & 0.5997 \\
Qwen-vl-Max & 0.6905 & 0.6995 & 0.5588 & 0.5588 & 0.5588 & 0.6480 & 0.6731 & 0.6667 & 0.6661 \\
Qwen-vl-plus & 0.837 & 0.8952 & 0.76 & 0.76 & 0.76 & 0.8194 & 0.8777 & 0.8506 & 0.8567 \\
\hline
\multicolumn{2}{l}{\textbf{Opened-Source VLMs}}\\
\hdashline
CogVLM-17B (Vicuna-v1.5-7B) & 0.9481 & 0.958 & 0.8467 & 0.8467 & 0.8467 & 0.9182 & 0.9431 & 0.9366  & 0.9334 \\
DeepSeek-VL-7B-chat & 0.9850 & 0.9669 & 0.9915 & 0.9915 & 0.9915 & 0.9784 & 0.9691 & 0.9882 & 0.9764 \\
InternLM-XComposer2-VL-7B (InternLM2) & 0.6536 & 0.5596 & 0.3789 & 0.3789 & 0.3789 & 0.5026 & 0.5447 & 0.5879 & 0.5456 \\
LLaVA-v1.5 (Vicuna-v1.5-7B) & 1.0000 & 0.9987 & 1.0000 & 1.0000 & 1.0000 & 1.0000 & 0.9983 & 1.0000  & 0.9992 \\
LLaVA-v1.5 (Vicuna-v1.5-13B) & 0.8802 & 0.8424 & 0.7981 & 0.7981 & 0.7981  & 0.8412 & 0.8209 & 0.8601 & 0.8367 \\
LLaVA-NeXT (Llama3-8B) & 0.3462 & 0.1864 & 0.3171 & 0.3171 & 0.3171 & 0.2452 & 0.2069 & 0.2690 & 0.2315 \\
mPLUG-Owl2 (LLaMA-7B) & 0.4545 & 0.2706 & 0.2966 & 0.2966 & 0.2966 & 0.3348 & 0.2914 & 0.3369& 0.3137 \\
Qwen-VL-Chat & 0.9621 & 0.9117 & 0.9697 & 0.9697 & 0.9697 & 0.9174 & 0.9174 & 0.9395 & 0.9340 \\
\hline
\end{tabular}
\end{table*}

\subsection{Appendix B: Details on LLM-assisted Evaluation for VQA}
\subsubsection{Settings for LLM Evaluation for VQA}

Our observations have revealed that some VLMs do not output answers in the specified format. For instance, in binary questions, responses might appear as ``True,'' ``The answer is true'', or ``It is true, because...''. To address this issue, after collecting responses from various VLMs for these binary and multiple-choice questions, we employed a \textit{LLM-Assisted Evaluation} method. This approach involves inputting the questions, correct answers, and VLM responses into a large language model (LLM) to assess the accuracy of the responses. Qwen- plus assisted in the evaluation of all models.

To mitigate the inherent variability of LLMs, where identical prompts can yield non-definitive responses, we employ a 5-round voting strategy. For each question-answer pair, we send the prompt defined in the template below five times and determined the correctness of the response based on the majority decision, selecting the outcome that occurred three times or more.

Although \textit{LLM-Assisted Evaluation} is highly efficient, it can occasionally produce errors, particularly when there is a significant discrepancy between the model’s output format and the standard answer. To mitigate this issue, we performed manual secondary reviews on all cases marked as incorrect by the LLM and corrected the results accordingly. Therefore, our evaluation process combines both manual and \texttt{LLM}-assisted methods to ensure the precision of the quantitative results.

\subsubsection{Prompt Templates for LLM Evaluation for VQA}
\begin{itemize}
    \item Prompt for true-or-false question evaluation:

    \textit{\#user: Given the question [true-or-false quesion], the correct answer is [correct answer], and the respondent's answer is [VLM's answer]. Determine if the respondent's answer is correct (1) or incorrect (1). If the answers match(both `True' or both `False'), output 1. Otherwise, output 0. Only return the result as a single digit.}

    \item  Prompt for multiple-choice question evaluation:

    \textit{\#user: Given the question [multiple-choice question and options, the correct answer is the option [correct answer]. The respondent's answer is [VLMs answer]. Determine if the respondent's answer is correct (1) or incorrect (0). If uncertain, also provide 0. Only return the result as a single digit.]}
\end{itemize}

\subsection{Appendix C: Details on LLM-assisted Evaluation for Descriptions}
\subsubsection{Settings for LLM Evaluation for Descriptions and Human Study}
This work focuses on how VLMs understand visual illusions, which often lead to challenges and inaccuracies in image interpretation. Therefore, we assess VLM performance by examining the accuracy of their descriptions, particularly whether they align with physical reality or human sensory perception.

Previous studies have shown that single-modal language models are effective for evaluating language tasks. After collecting open-ended responses from the VLMs, we use advanced LLM to quantitatively assess multimodal description tasks. Specifically, both the model’s output and the golden description are input into the LLM, which compares the two to identify significant conflicts. Preciseness is scored on a scale of [0, 1, 2]. 
\begin{itemize}
    \item \textbf{Semantic Content Description: }
    
    2 indicates no conflict;
    
    1 indicates minor conflict;
    
    0 indicates a significant conflict.

    \item \textbf{Visual Illusion Description: }
    
    2 is given when there is no significant conflict; 
    
    1 is given when the existence of the illusion is correctly described, but there are minor conflicts in the explanation of its cause; 
    
    0 is given when the existence of the illusion is incorrectly described, or there are significant conflicts in the explanation of its cause. 
\end{itemize}
Qwen-plus assisted in the evaluation of all models.

To address the inherent variability of LLMs, where identical prompts can yield non-definitive responses, we employed a 5-round averaging strategy. For each descriptive output generated by the VLM, we sent the prompt defined below five times and averaged the scores to determine the final score for that description. This method effectively mitigates the inherent variability of LLMs.

Additionally, we randomly selected 200 images, proportionate to the image categories, for a human study sample. We recruited two human evaluators, each of whom assessed the descriptions generated by GPT-4o, Gemini-pro-1.5, and Qwen-vl-plus for 100 images. The Spearman Correlation Coefficients (SRCC) between the human scores and LLM scores were 0.9055 and 0.9246, respectively. These results strongly validate the effectiveness and accuracy of the LLM-assisted method for evaluating descriptions.

\subsubsection{Prompt Templates for LLM Evaluation for Description}
\begin{itemize}
    \item \textbf{Prompt for semantic content description evaluation:}
    
    \textit{\# user: Respondent description is `[VLM description]', reference description is `[Golden description]'. Evaluate if there is a conflict between the image contents in the respondent's answer and the reference answer. Rate as follows: 2: No conflict. 1: Minor conflict, less controversial than the reference. 0: Clear conflict, more controversial than the reference. Only focus on conflicts between the descriptions. Differences in detail or omitted information are not considered conflicts. Just answer the rate number, do not output any other word.}

    \item \textbf{Prompt for illusion description evaluation:}
    
    \textit{\# user: Respondent description is `[VLM description]', reference description is `[Golden description]'. Evaluate if there is a conflict between the existence and causal information of illusions in the respondent's description and the reference description. Rate as follows: 2: No conflict. 1: no conflict in the existence of illusions, but there are conflicts in the causes of illusions. 0: Clear conflict, or conflict in the existence of illusion. Only focus on conflicts between the descriptions. Differences in detail or omitted information are not considered conflicts. Just answer the rate number, do not output any other word.}
\end{itemize}

\begin{table*}[htb]
\centering
\fontsize{8pt}{9.2pt}\selectfont
\belowrulesep=0pt
\aboverulesep=0pt
\caption{Detailed performance of VLMs on IllusionBench+ description task across different image categories and difficulty levels.}
\label{description_detail}
\begin{tabular}{c|c|ccccc|ccc|c}
\hline
\multirow{2}{*}{\makecell{\textbf{VLMs}}}&\multirow{2}{*}{\makecell{\textbf{Question Type}}} & \multicolumn{5}{c}{\textbf{Image Category}} & \multicolumn{3}{c}{\textbf{Difficulty Rating}}  & \multirow{2}{*}{\makecell{\textbf{All}}} \\
\cmidrule(lr){3-10} 
& & Classical & Real Scene & No Illusion & Ishihara & Trap & Easy & Neutral &Hard  \\
\hline
\textbf{\rule{0pt}{2ex}Closed-Source VLMs\rule[-1ex]{0pt}{0pt}} &&&&&&&&&&\\
\hdashline
\multirow{4}{*}{GPT-4o} & Semantic Content &1.2805 & 1.2269 & 1.4939 & 1.4321 & 0.3333 & 1.4317 & 1.2196 & 1.1892 & 1.2872 \\
& Visual Illusion & 1.0793 & 1.0673 & 0.6694 & 0.8462 & 0.0000 & 0.8634 & 1.0675 & 0.8604 & 0.9522 \\
& Step-by-Step&  1.3354 &1.3815 & 1.4041 & 1.2308 & 0.5556 & 1.4426 & 1.3652 & 1.2207 & 1.3616 \\
& All & 1.2317 & 1.2253 & 1.1891 & 1.1667 & 0.2963 & 1.2459 & 1.2175 & 1.0901 & 1.2004 \\
\hdashline
\multirow{4}{*}{Gemini-pro-1.5} & Semantic Content & 0.8962 & 0.9294 & 1.3633 & 1.2308 & 0.5263 & 1.2822 & 0.8918 & 0.8829 & 1.0257 \\
& Visual Illusion  & 1.0244 & 1.0790 & 1.3244 & 0.7692 & 0.4211 & 1.2795 & 1.0823 & 0.8784 & 1.1077\\
& Step-by-Step & 1.0793 & 1.2286 & 1.2939 & 0.8846 & 0.3158 & 1.3096 & 1.2121 & 0.9730 & 1.1954\\
& All & 1.0000 & 1.0790 & 1.3265 & 0.9615 & 0.4211 & 1.2904 & 1.0620 & 0.9144 & 1.1096\\
\hdashline
\multirow{4}{*}{Qwen-vl-Max} & Semantic Content & 0.6768 & 0.6745 & 1.1020 & 0.2308 & 0.3158 & 0.9863 & 0.6429 & 0.6171 & 0.7571  \\
& Visual Illusion  & 0.6748 & 0.7315 & 1.4857 & 0.9615 & 0.0000 & 1.1967 & 0.8026 & 0.5721 & 0.8913\\
& Step-by-Step & 0.8720 & 0.8809 & 1.0653 & 0.6923 & 0.1053 & 1.0109 & 0.8853 & 0.7658 & 0.9038\\
& All & 0.7413 & 0.7623 & 1.2177 & 0.6282 & 0.1404 & 1.0647 & 0.7769 & 0.6517 & 0.8507 \\
\hdashline
\multirow{4}{*}{Qwen-vl-plus}  & Semantic Content & 0.7134 & 0.6655 & 1.1903 & 0.8077 & 0.2222 & 1.019 & 0.6848 & 0.6396 & 0.7924 \\
& Visual Illusion  & 0.6829 & 0.7049 & 0.8097 & 0.3462 & 0.0000 & 0.7174 & 0.7495 & 0.5928 & 0.7052\\
& Step-by-Step & 0.7500 & 0.7361 & 1.0607 & 0.4615 & 0.4444 & 0.9620 & 0.7283 & 0.6937 & 0.8029\\
& All & 0.7154 & 0.7022 & 1.0202 & 0.5385 & 0.2222 & 0.8995 & 0.7208 & 0.6421 & 0.7668 \\
\hline
\textbf{\rule{0pt}{2.5ex}Opened-Source VLMs\rule[-1ex]{0pt}{0pt}} &&&&&&&&&&\\
\hdashline
\multirow{4}{*}{\makecell{CogVLM-17B\\(Vicuna-v1.5-7B)}}& Semantic Content & 0.7866 & 0.8208 & 1.2490 & 0.5000 & 0.4211 & 1.1776 & 0.7646 & 0.7252 & 0.9001  \\
& Visual Illusion  & 0.5976 & 0.6951 & 0.9592 & 0.3462 & 0.1053 & 0.8880 & 0.6760 & 0.5450 & 0.7222\\
& Step-by-Step & 0.7012 & 0.8442 & 1.1020 & 0.3077 & 0.0000 & 1.0383 & 0.7927 & 0.6757 & 0.8535\\
& All & 0.6951 & 0.7867 & 1.1034 & 0.3846 & 0.1754 & 1.0346 & 0.7444 & 0.6486 & 0.8252 \\
\hdashline
\multirow{4}{*}{DeepSeek-VL-7B-chat}& Semantic Content & 0.7707 & 0.6215 & 1.1602 & 0.3600 & 0.2105 & 1.0259 & 0.6577 & 0.5096 & 0.7550  \\
& Visual Illusion  & 0.5556 & 0.6092 & 1.1982 & 0.4400 & 0.3333 & 1.0175 & 0.6103 & 0.4900 & 0.7293\\
& Step-by-Step & 0.7351 & 0.7610 & 1.1045 & 0.1923 & 0.0556 & 0.9881 & 0.7642 & 0.5930 & 0.8071\\
& All & 0.6876 & 0.6632 & 1.1549 & 0.3289 & 0.2000 & 1.0107 & 0.6770 & 0.5305 & 0.7636 \\
\hdashline
\multirow{4}{*}{\makecell{InternLM-XComposer2\\-VL-7B (InternLM2)}}& Semantic Content & 0.5915 & 0.6365 & 1.2204 & 0.1538 & 0.0526 & 1.0628 & 0.6004 & 0.5135 & 0.7431  \\
& Visual Illusion  & 0.6380 & 0.6683 & 0.7306 & 0.0769 & 0.2105 & 0.7104 & 0.6364 & 0.6036 & 0.6552\\
& Step-by-Step & 0.5854 & 0.6298 & 1.2041 & 0.0769 & 0.1053 & 1.0410 & 0.5810 & 0.5450 & 0.7336\\
& All & 0.6049 & 0.6449 & 1.0517 & 0.1026 & 0.1228 & 1.9381 & 0.6059 & 0.5541 & 0.7107 \\
\hdashline
\multirow{4}{*}{\makecell{LLaVA-v1.5\\(Vicuna-v1.5-7B)}}& Semantic Content & 0.3841 & 0.3735 & 0.8939 & 0.0000 & 0.0526 & 0.7104 & 0.3629 & 0.3514 & 0.4814  \\
& Visual Illusion  & 0.4268 & 0.4355 & 0.0816 & 0.0000 & 0.0000 & 0.1995 & 0.4212 & 0.3694 & 0.333\\
& Step-by-Step & 0.3476 & 0.4322 & 0.5061 & 0.0000 & 0.0000 & 0.5000 & 0.3801 & 0.3604 & 0.4177\\
& All & 0.3862 & 0.4137 & 0.4939 & 0.0000 & 0.0175 & 0.4699 & 0.3880 & 0.3604 & 0.4177 \\
\hdashline
\multirow{4}{*}{\makecell{LLaVA-v1.5\\(Vicuna-v1.5-13B)}}& Semantic Content & 0.4207 & 0.3786 & 1.0245 & 0.3846 & 0.0000 & 0.8033 & 0.3823 & 0.3829 & 0.5290  \\
& Visual Illusion  & 0.4939 & 0.4420 & 0.1475 & 0.0000 & 0.0000 & 0.2521 & 0.4394 & 0.3846 & 0.3626\\
& Step-by-Step & 0.5061 & 0.4104 & 0.2735 & 0.0769 & 0.0000 & 0.3033 & 0.4276 & 0.3964 & 0.3777\\
& All & 0.4736 & 0.4103 & 0.4823 & 0.1538 & 0.0000 & 0.4531 & 0.4164 & 0.388 & 0.4232 \\
\hdashline
\multirow{4}{*}{\makecell{LLaVA-NeXT\\(Llama3-8B)}}& Semantic Content & 0.5122 & 0.6114 & 1.2898 & 0.1154 & 0.3158 & 1.0219 & 0.6069 & 0.5360 & 0.7364  \\
& Visual Illusion  & 0.4939 & 0.6968 & 1.0245 & 0.0385 & 0.2105& 0.8962 & 0.6717 & 0.5135 & 0.7165\\
& Step-by-Step & 0.6951 & 0.8844 & 0.8857 & 0.0000 & 0.1053 & 0.8743 & 0.8337 & 0.6982 & 0.8192\\
& All & 0.5671 & 0.7309 & 1.0667 & 0.0513 & 0.2105 & 0.9308 & 0.7041 & 0.5826 & 0.7574 \\
\hdashline
\multirow{4}{*}{\makecell{mPLUG-Owl2\\(LLaMA-7B)}}& Semantic Content & 0.5122 & 0.4707 & 1.0449 & 0.1538 & 0.1579 & 0.8743 & 0.4838 & 0.3784 & 0.5975  \\
& Visual Illusion  & 0.4540 & 0.4966 & 0.1347 & 0.0000 & 0.0000& 0.2747 & 0.4892 & 0.3439 & 0.3840\\
& Step-by-Step & 0.5122 & 0.5946 & 0.5184 & 0.0000 & 0.0000 & 0.5710 & 0.5508 & 0.4595 & 0.5385\\
& All & 0.4929 & 0.5207 & 1.5660 & 0.0513 & 0.0526 & 0.5739 & 0.5079 & 0.3940 & 0.5068 \\
\hdashline
\multirow{4}{*}{Qwen-VL-Chat}& Semantic Content & 0.6524 & 0.6399 & 1.1020 & 0.3077 & 0.2105 & 0.9973 & 0.6199 & 0.5360 & 0.7336  \\
& Visual Illusion  & 0.5488 & 0.5731 & 0.7633 & 0.0000 & 0.0000 & 0.7240 & 0.5553 & 0.4369 & 0.5891\\
& Step-by-Step & 0.6890 & 0.7320 & 0.9510 & 0.0769 & 0.1053 & 0.9235 & 0.6674 & 0.6306 & 0.7488\\
& All & 0.6301 & 0.6484 & 0.9388 & 0.1282 & 0.1053 & 0.8816 & 0.6143 & 0.5345 & 0.6906 \\
\hline
\end{tabular}
\end{table*}

\newpage

\subsection{Appendix D: Extended Experiment Result on IllusionBench+ Description Task}
\subsubsection{Detailed Performance Result for Description Task}
\begin{figure*}[htb]
    \centering
    \begin{subfigure}[b]{0.9\textwidth}
        \hspace*{0cm}
        \includegraphics[width=\textwidth]{ 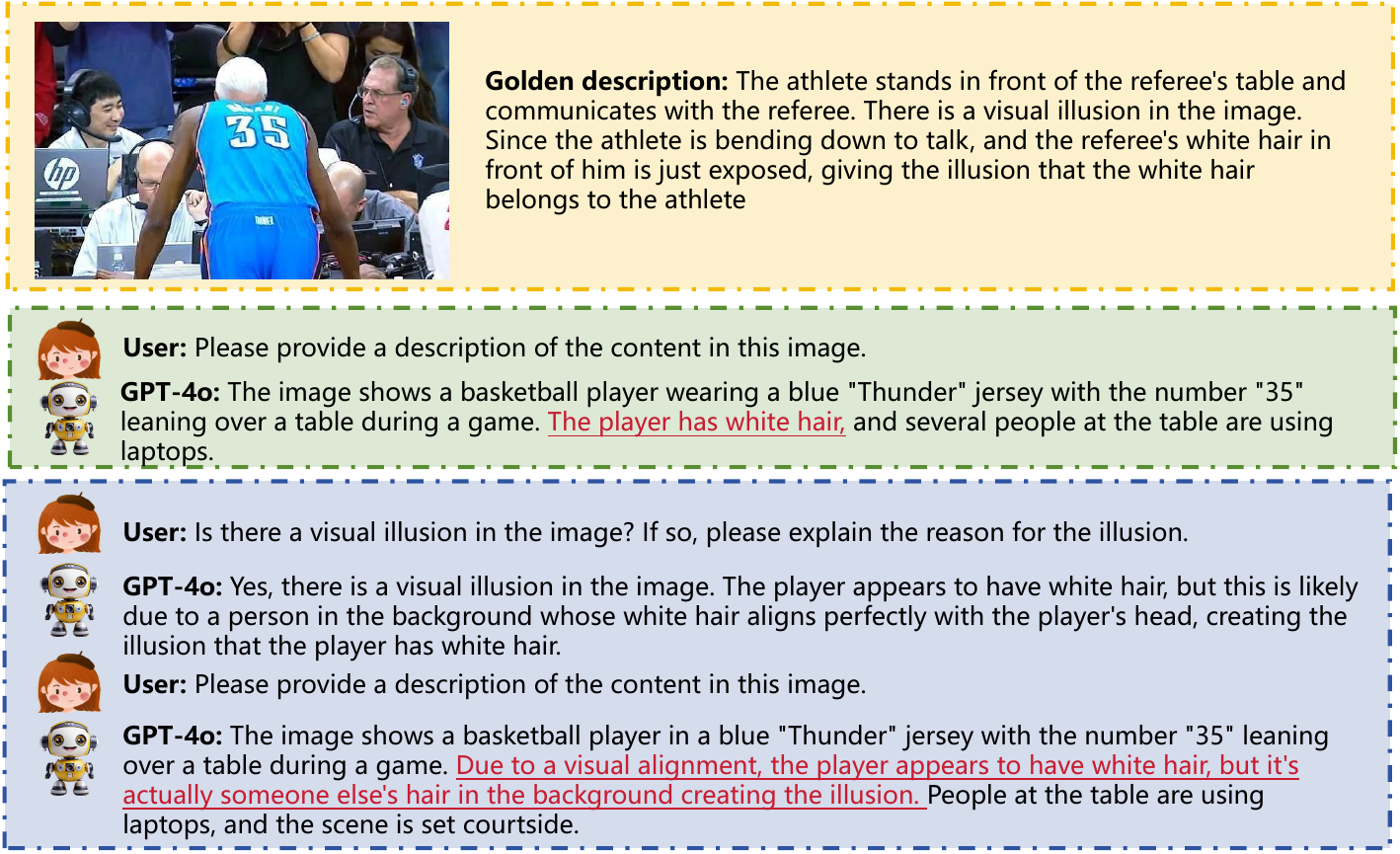} 
    \end{subfigure}  
    \\
    \begin{subfigure}[b]{0.905\textwidth}
        \hspace*{0cm}
        \includegraphics[width=\textwidth]{ 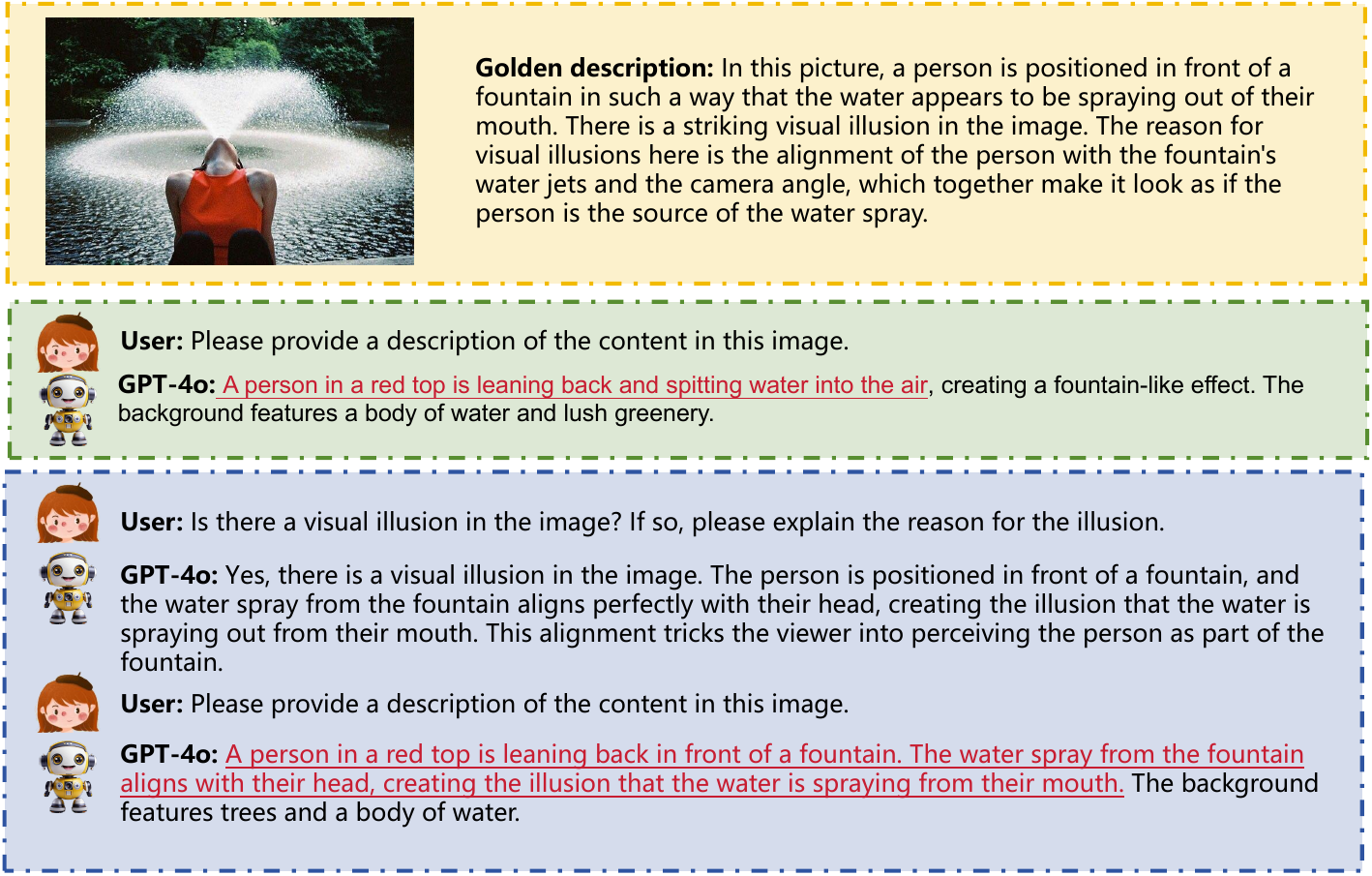} 
    \end{subfigure}  
    \caption{Example of VLM in describing real scene illusion images.}
    \label{real_scene_example}
\end{figure*}

\begin{figure*}[htb]
    \centering
    \includegraphics[width=0.9 \textwidth]{ 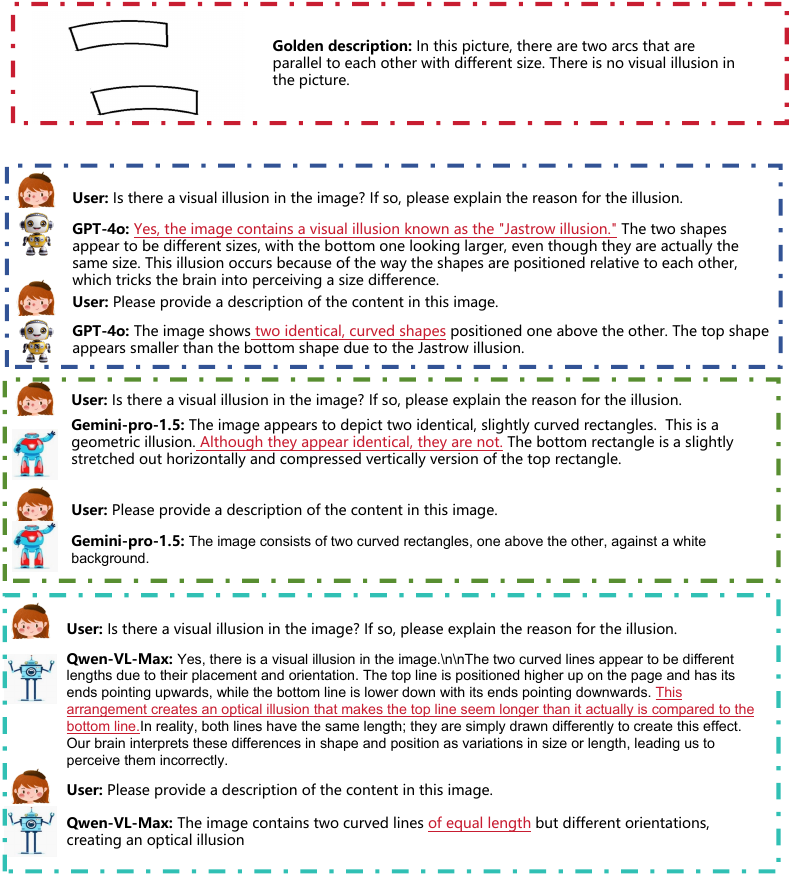} 
    \caption{Example of VLMs in describing trap illusion images.}
    \label{trap_example}
\end{figure*}
The detailed performance results for VLMs on the open-ended description task are shown in Table \ref{description_detail}, revealing several key insights:

\begin{itemize}
    \item \textbf{There is a significant gap between the description capabilities of VLMs for illusion images and those of humans.} Only the most advanced models, such as GPT-4o and Gemini-pro-1.5, achieved an average description score exceeding 1. Additionally, for most VLMs, the preciseness of their descriptions across various questions declines as the human-assigned difficulty scores increase. This suggests that as the complexity of the tasks rises, these VLMs struggle to maintain accuracy in their generated descriptions, indicating that their current abilities to comprehend and process complex contexts or problems still need improvement.

    \item \textbf{The presence of visual illusions indeed impacts VLMs' ability to describe images precisely.} Most models perform best when describing the semantic content of images in the ``no illusion" category, suggesting that visual illusions pose a significant challenge to VLMs' understanding.
    
    \item \textbf{The bias of VLMs towards answering ``true" affects the effectiveness of the step-by-step approach.} The results show that higher performance under the step-by-step strategy compared to directly performing the semantic content description task typically occurs in the ``classical illusion" and ``real scene illusion" categories. However, in the ``no illusion" category, the step-by-step approach almost invariably leads to a decline in descriptive performance. This may be because VLMs tend to answer ``true" when asked about the presence of illusions, causing interference in the subsequent semantic description of the image, thereby reducing the preciseness of their descriptions.

\end{itemize}

\subsubsection{Examples for Description Tasks in IllusionBench+}

\begin{itemize}
    \item \textbf{The step-by-step strategy can effectively improve the preciseness of VLMs in image semantic content description.}\\
    Fig. \ref{real_scene_example} presents two examples from the``real scene illusion" category where GPT-4o misinterpreted the images due to the illusions present. However, by employing the step-by-step strategy, first inquiring about the existence and cause of the visual illusion, the model's potential can be harnessed, leading to more precise semantic descriptions of the images.

    \item \textbf{The classical cognitive illusion is no longer sufficient to test the alignment of VLMs with human vision. }\\
    Fig. \ref{trap_example} displays the performance of various state-of-the-art closed-source VLMs on a trap illusion description task. The example in Fig. \ref{trap_example} is an edited version of the Jastrow illusion. The Jastrow illusion is a well-known optical illusion where two identical curved shapes are placed side by side, with one rotated, making the lower one appear larger even though they are actually the same size. In our edited version, however, the lower curved shape is indeed larger than the upper one. As shown in Fig. \ref{trap_example}, only Gemini-pro-1.5 correctly identified that the two curved shapes differ in size, while GPT-4o and Qwen-VL-Max both responded according to the standard Jastrow illusion conclusion, leading to a perception error. This phenomenon is consistent with the findings in Table. \ref{description_detail}, where GPT-4o demonstrates the highest accuracy in describing classical illusions but performs very poorly in the trap illusion category, answering all questions incorrectly in the illusion description task. This suggests that the most advanced VLMs have fully learned the knowledge embedded in classical cognitive illusion images, to the extent that they experience hallucinations when presented with images that are similar in pattern but differ in physical reality. Consequently, classical cognitive illusion images are no longer sufficient to achieve the original goal of testing the alignment between VLMs and human vision, highlighting the need for a more comprehensive dataset like IllusionBench+ to address this issue.
\end{itemize}

\end{document}